\documentclass{article}

\usepackage{arxiv}

\usepackage[utf8]{inputenc} 
\usepackage[T1]{fontenc}    
\usepackage{hyperref}       
\usepackage{url}            
\usepackage{booktabs}       
\usepackage{amsfonts}       
\usepackage{nicefrac}       
\usepackage{microtype}      
\usepackage{lipsum}
\usepackage{amsmath,amssymb,amsfonts, algorithm2e}
\usepackage{algorithmic}
\usepackage{graphicx}
\usepackage{textcomp}
\usepackage{xcolor}
\newcommand{\etal}{\textit{et al.}}
\usepackage{multirow}

\title{PCRP: Unsupervised Point Cloud Object \\ Retrieval and Pose Estimation}

\author{
  Pranav Kadam \\
  Media Communications Lab \\
  University of Southern California\\
  Los Angeles, CA, USA \\
  \texttt{pranavka@usc.edu} \\
  
  \And
 Qingyang Zhou \\
  Media Communications Lab\\
  University of Southern California\\
  Los Angeles, CA, USA \\
  \texttt{qzhou776@usc.edu} \\
  
   \And
 Shan Liu\thanks{This work was supported by Tencent Media Lab.} \\
  Tencent Media Lab\\
  Tencent America\\
  Palo Alto, CA, USA \\
  \texttt{shanl@tencent.com} \\
  
  \And
 C.-C. Jay Kuo \\
  Media Communications Lab\\
  University of Southern California\\
  Los Angeles, CA, USA \\
  \texttt{cckuo@sipi.usc.edu} \\
}

\begin{document}
\maketitle

\begin{abstract}
An unsupervised point cloud object retrieval and pose estimation method,
called PCRP, is proposed in this work. It is assumed that there exists a
gallery point cloud set that contains point cloud objects with given
pose orientation information. PCRP attempts to register the unknown
point cloud object with those in the gallery set so as to achieve
content-based object retrieval and pose estimation jointly, where the
point cloud registration task is built upon an enhanced version of the
unsupervised R-PointHop method. Experiments on the ModelNet40 dataset
demonstrate the superior performance of PCRP in comparison with
traditional and learning based methods. 
\end{abstract}


\section{Introduction}\label{sec:intro}

Content-based point cloud object retrieval and category-level point
cloud object pose estimation are two important tasks of point cloud
processing. For the former, one can find similar objects from the
gallery set, which can provide more information about the unknown
object.  For the latter, the goal is to estimate the 6-DOF pose of a 3D
object comprising of rotation $(R \in SO(3))$ and translation $(t \in
\mathbb{R}^3)$ , with respect to a chosen reference.  The pose
information can facilitate downstream tasks such as object grasping,
obstacle avoidance and path planning for robotics.  In a typical scene
understanding problem using data from range sensors or a depth camera,
this problem would arise after a 3D detection algorithm has successfully
localized and labeled the objects present in the point cloud scan. 
An unsupervised point cloud object retrieval and pose estimation method,
called PCRP, is proposed in this work. 

\begin{figure}[htb]
\centerline{\includegraphics[width=4in]{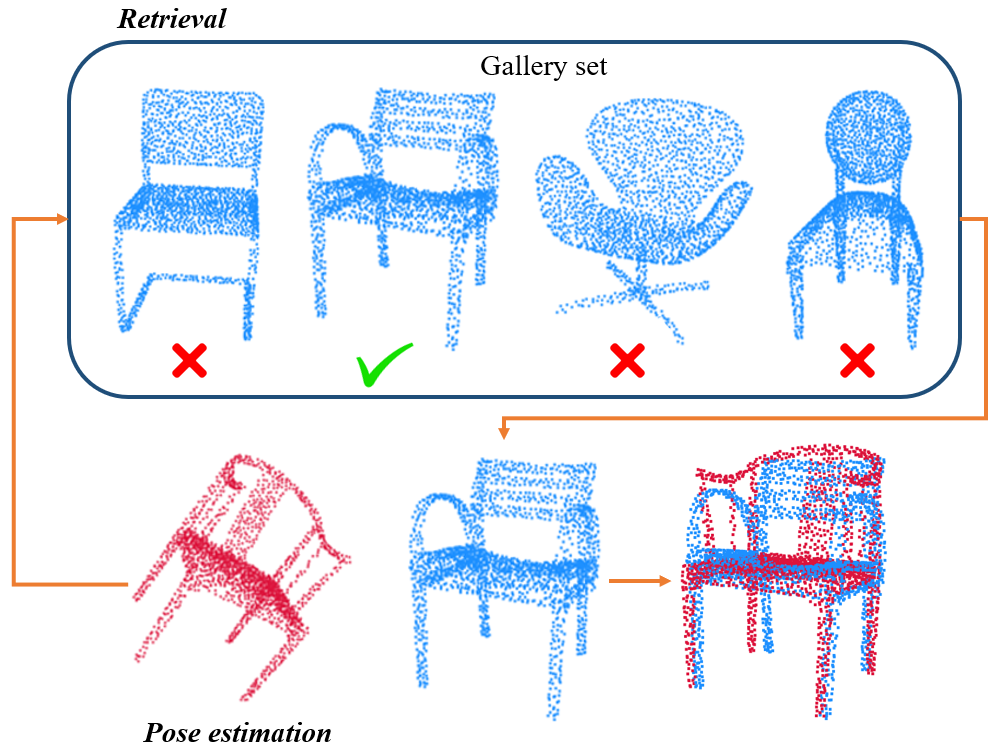}}
\caption{Summary of the proposed PCRP method. First, a similar object to the
input query object (in red) is retrieved from the gallery set (top row).
Then, the query object is registered with the retrieved object (bottom
row) to estimate its pose.}\label{fig:summary}
\end{figure}

The R-PointHop method was recently proposed by Kadam \etal
\cite{kadam2021r} for point cloud registration. Although being
unsupervised, it offers competitive performance with respect to other
supervised learning methods.  In this paper, we extend R-PointHop to the
context of point cloud object retrieval and pose estimation against a
gallery point cloud set, which contains point cloud objects with known
pose orientation information. Point cloud retrieval has been researched
for quite some time in terms of retrieving a similar object from a
database or aggregating local feature descriptors for recognizing places
in the wild. Yet, retrieving objects with pose variations is less
investigated. Here, we show how R-PointHop features can be reused to
retrieve a similar point cloud object. Registration of two similar
objects (potentially with partial overlap), which was the focus of
R-PointHop, has been widely studied. Although being a related problem,
estimating the pose of a single object addressed in this work is less
explored. We analyze several bottlenecks of R-PointHop and propose
modifications to enhance its performance for pose estimation. 

Built upon enhanced R-PointHop, PCRP registers the unknown point cloud
object with those in the gallery set to achieve content-based object
retrieval and pose estimation jointly.  As shown in Fig.
\ref{fig:summary}, PCRP consists of ``object retrieval" and ``pose
estimation" two functions. For object retrieval, it first aggregates the
pointwise features learned from R-PointHop based on VLAD (Vector of
Locally Aggregated Descriptors) \cite{jegou2010aggregating} to obtain a
global point cloud feature vector and then use it to retrieve a similar
pre-aligned object from the gallery set. For pose estimation, the 6-DOF
pose of the query object is found by registering it with the retrieved
object. Experiments on the ModelNet40 dataset demonstrate the superior
performance of PCRP in comparison with traditional and learning based
methods. 

This work has two main contributions. First, we extend R-PointHop, which
was originally designed for point cloud registration, to object
retrieval and pose estimation. We show how features derived from
R-PointHop can be aggregated to yield a global feature descriptor for
object retrieval and reuse point features for pose estimation. Second, we
propose ways to modify the attribute representation in R-PointHop and
make it more general. As a result, any traditional point local
descriptors, such as FPFH \cite{rusu2009fast} and SHOT
\cite{tombari2010unique}, can be adopted by R-PointHop.  The rest of the
paper is organized as follows. Related previous work is reviewed in Sec.
\ref{sec:review}. The PCRP method is proposed in Sec.  \ref{sec:method}.
Experimental results are shown in Sec.  \ref{sec:results}. Finally,
concluding remarks are given in Sec.  \ref{sec:conclusion}. 

\section{Review of Related Work}\label{sec:review}

Point cloud processing includes classification, registration,
segmentation, retrieval, pose estimation, etc. There has been major
advancement in point cloud processing due to deep learning.  PointNet
\cite{qi2017pointnet} employed an MLP-based permutation invariant
network for point cloud classification and segmentation.  It is followed
by a series of work \cite{qi2017pointnet++, wang2019dynamic,
li2018pointcnn}. Besides permutation invariance, features invariant to
point cloud rotation \cite{deng2018ppf, gojcic2019perfect} are desirable
for applications such as data association, registration and
classification of unaligned objects.  Learning based global registration
methods \cite{aoki2019pointnetlk, wang2019deep, kadam2021r} outperform
traditional handcrafted descriptors such as FPFH \cite{rusu2009fast} and
SHOT \cite{tombari2010unique} in registration performance and ICP-based
\cite{besl1992method} methods in object registration.  Recently, point
cloud equivariant networks \cite{deng2021vector, chen2021equivariant,
li2021leveraging} show impressive performance for object pose
estimation, retrieval and classification under different poses.
Furthermore, the learning based methods can be successfully applied to
complex indoor scene registration and visual odometry estimation.

\begin{figure}[htb]
\centerline{\includegraphics[width=5.5in]{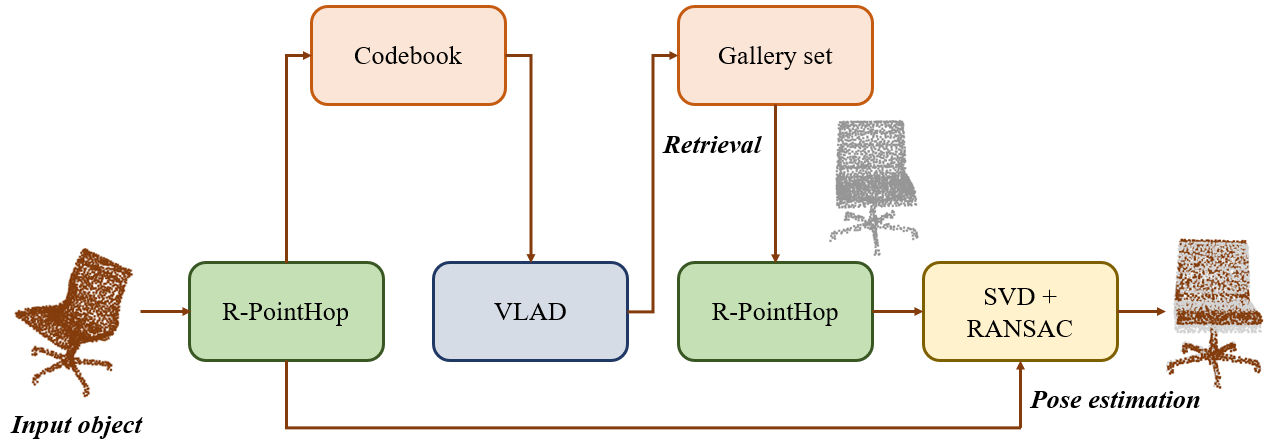}}
\caption{An overview of the proposed PCRP method: 1) pointwise feature
extraction from the input point cloud using R-PointHop, 2) aggregation
of point features into a global VLAD representation and retrieval of
similar objects from the gallery set based on similarity of VLAD
representations, and 3) the pose of the input is estimated by
registering the query object with the retrieved object.}
\label{fig:blockdiagram}
\end{figure}

As an alternative to deep learning, successive subspace learning (SSL)
has been proposed for point cloud processing. Its potential was
demonstrated in PointHop \cite{zhang2020pointhop}, PointHop++
\cite{zhang2020pointhop++}, SPA \cite{kadam2020unsupervised}, R-PointHop
\cite{kadam2021r} and GPCO \cite{kadam2021gpco}. The unsupervised
feature learning in SSL consists of attribute construction, neighborhood
expansion, and dimensionality reduction using the Saab transform.
R-PointHop extracts a local reference frame for every point and 
builds the local feature by aligning neighborhood points to it. The
neighborhood grows successively and short-, mid-, and long-range point
information is captured. Later, a set of point correspondences are found
and the 3D transformation is estimated using singular value
decomposition of the covariance matrix of corresponding points. 

One limitation of R-PointHop and other exemplary deep networks for
pair-wise registration \cite{aoki2019pointnetlk, wang2019deep,
wang2019prnet} is the assumption that a pre-aligned reference object is
available, which is the same instance of the input object whose pose is
to be estimated.  However, such an assumption may not hold in practice
and, as a result, the pose estimation problem cannot be solved by
registration.  One way to avoid this difficulty is to retrieve a similar
pre-aligned object from a gallery set first and then estimate the object
pose by registering it with the retrieved object as done in CORSAIR
\cite{zhao2021corsair}.  CORSAIR uses the bottleneck layer
representation of a registration network to train another network for
retrieval using metric learning. It is worthwhile to mention that some
pose estimation methods are reference object free
\cite{li2021leveraging, chen2021equivariant}. They adopt point cloud
equivariant networks as the backbone and do not use object retrieval for
pose estimation. 

Point cloud retrieval has been studied extensively due to its rich
applications. Noteworthy work includes 3D ShapeNets \cite{wu20153d} for
shape retrieval and PointNetVLAD \cite{uy2018pointnetvlad} for place
recognition. Our retrieval idea is inspired by PointNetVLAD, which
aggregates pointwise features from PointNet using the VLAD feature
\cite{jegou2010aggregating}. Their VLAD feature is obtained by taking a
weighted average of point features, where weights are jointly trained
with the PointNet network under supervised learning. In contrast, our
R-PointHop feature is learned without supervision.  We should emphasize
that most retrieval methods assume that the query object and objects
from the gallery set are pre-aligned. Yet, in the context of joint
object retrieval and pose estimation, this assumption does not hold. It
is essential to develop a retrieval method, where the query object can
possess any arbitrary rotation. Our work is uniquely positioned in
addressing two challenges at one shot: 1) pose estimation without
identical object instances, and 2) object retrieval without
pre-alignment. Furthermore, we do not use deep leaning but successive
subspace learning. 

\section{Proposed PCRP Method}\label{sec:method}

As shown in Fig. \ref{fig:blockdiagram}, the PCRP method consists of the
following three stages. First, features of every point in the input
point cloud are extracted using R-PointHop. Second, features of all
points are aggregated into a global descriptor using the VLAD method.
The nearest neighbor search is then used to retrieve a similar
pre-aligned object from the gallery set. Finally, the input object is
registered to the retrieved object to obtain the 6-DOF pose. Each of
them is elaborated below. 

\subsection{Feature Extraction}

In the R-PointHop originally proposed in \cite{kadam2021r}, point
attributes are constructed by partitioning the 3D space around a point
into eight octants using three orthogonal directions given by the local
reference frame. The mean of points in each octant is concatenated to
get a 24D attribute vector. Yet, we observe that the 24D attribute
vector is sensitive to noise and unable to capture complex local surface
patterns for distinction. A modified version that appends point
coordinates with eigen features was used for indoor scene registration
and odometry \cite{kadam2021gpco}.  Actually, histogram-based point
descriptors such as SHOT (Signature Histogram of Orientations)
\cite{tombari2010unique} and FPFH (Fast Point Feature Histogram)
\cite{rusu2009fast} have been widely used to describe the local surface
geometry. We may leverage them as well. One drawback of histogram-based
descriptors is that they cannot capture the far-distance information
since they have a single scale only. 

We propose to integrate histogram-based local descriptors with
R-PointHop, which has a multi-scale representation capability, to get a
new descriptor. Specifically, we replace the octant-based
mean-coordinates attributes in the original R-PointHop with the FPFH
descriptor in the first hop. The rest is kept the same as the original
R-PointHop. That is, the Saab transform is used to get the first-hop
spectral representation, and subsequent hops still involve attribute
construction by partitioning the 3D space into eight octants and
getting 8D attributes for each spectral component. We should emphasize
that FPFH is invariant to rotations so that the rotation-invariant
property of R-PointHop is preserved without any other adjustment.  The
output of the first-hop stemming from FPFH is more powerful than the
original R-PointHop design. We give the new descriptor a name -
FR-PointHop.  This modification enriches R-PointHop since it can take
any local rotation-invariant feature representation and generalize it to
a multi-scale descriptor using the standard R-PointHop pipeline. 

\subsection{Feature Aggregation}

For an input point cloud, we extract features of each point using
FR-PointHop. Features of all points need to be aggregated to yield a
global descriptor for object retrieval.  One choice of aggregation is to
use global max/mean pooling. It has been widely adopted in point cloud
classification. However, global pooling is not a good choice since point
features obtained by FR-PointHop only cover the information of a local
neighborhood. In the retrieval literature, Bag of Words (BoW) and
vectors of locally aggregated descriptors (VLAD)
\cite{jegou2010aggregating} are popular methods in aggregating local
features such as SIFT. Here, we adopt VLAD \cite{jegou2010aggregating}
to aggregate point descriptors obtained by R-PointHop to yield a global
feature vector that is suitable for retrieval. The global feature
aggregation process is stated below. 

The first step is to generate a codebook of $k$ codewords of $d$
dimensions, where $d$ is the feature dimension. The k-means clustering
algorithm is used to achieve this objective. Learned point features from
the training data are used to form clusters, whose centroids are
computed.  The $k$ centroids represent $k$ codewords in the codebook.
Given an input point cloud, its global VLAD feature vector is calculated
as follows. The feature vector of each point is first assigned to the
nearest codeword based on the shortest distance criterion. Next, for
each point, the difference between its descriptor and the assigned
codeword is calculated, which represents the error vector in the feature
space. Then, the differences with respect to the same codeword are added
together.  Finally, error sums of all codewords are concatenated to get
the VLAD feature.  For a $d$-dimensional feature vector, the VLAD
feature is of dimension $k \times d$, where $k$ is the codeword number. 

In the training process, we use the generated codebook to pre-compute
the VLAD features of point cloud objects in the gallery set.  They are
stored along with the codebook.  In the inference stage, the query
object is first passed through FR-PointHop to extract pointwise
features.  Then, its VLAD feature is calculated using the same codebook
and compared with the VLAD features of point cloud objects from the
gallery set. The nearest neighbor search is used to retrieve the best
matching point cloud. It is worthwhile to mention that, due to the
rotation invariant nature of FR-PointHop, point features and, hence,
VLAD features are invariant with object's pose, which facilities
retrieval in presence of pose variations. 

\subsection{Pose Estimation}

Once an object from the gallery set is retrieved, the next step is to
register the query and the retrieved objects. The process closely
resembles the registration task of R-PointHop with additional aid of the
object symmetry information.  The pointwise features extracted from the
query object for retrieval are reused here. For objects in the gallery
set, we only store their VLAD descriptors rather than their pointwise
features due to the high memory cost of the latter. Yet, pointwise
features of the retrieved object can be computed again using FR-PointHop
at run time.  The cost is manageable since it is done for one retrieved
object. Afterwards, corresponding points between query and retrieved
objects are found using the nearest neighbor criterion in the feature
space. 

We exploit the object symmetry information to limit the correspondence
search region. 3D objects often possess different forms of symmetry. For
example, chair objects have a planar symmetry.  To avoid mismatched
correspondences arising due to object symmetry, correspondences are
constrained to be among disjoint sets of points.  For every object, we
divide its points into two disjoint sets using its principal components.

\begin{figure}[htb]
\centerline{\includegraphics[width=4.5in]{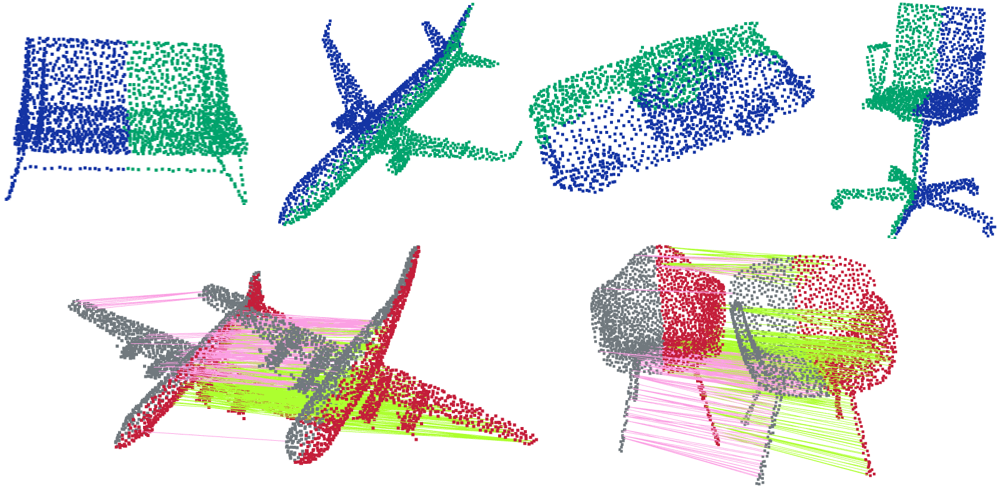}}
\caption{(Top) Illustration of partitioning of point clouds into two
symmetrical parts and (Bottom) point correspondences between symmetric
parts.}\label{fig:symmetry}
\end{figure}

First, we calculate 1D moments of the points projected along each
principal direction about the origin. Then, we take the absolute
difference between sum of moments of positive coordinates and negative
coordinates. Then, the principal component with the least absolute
difference is used to divide the points in disjoint sets. The main
intuition is to find an axis along which the projection of points is
most symmetric. The search space for this axis is restricted to object's
principal components for simplicity. Yet, it yields reasonable results
as seen from top row of Fig. \ref{fig:symmetry}. Afterwards, the point
correspondences are found in each disjoint set (see the bottom row of
Fig.  \ref{fig:symmetry}) and then concatenated. 

Once the point correspondence is built, we use the orthogonal procrustes
method \cite{schonemann1966generalized}, which is based on the singular
value decomposition (SVD) of the data covariance matrix, to estimate the
rotation and translation that aligns the query object to the retrieved
object.  This part is identical with that in R-PointHop.  The
transformation is optimal in the sense that it achieves the least sum of
squared errors between matching points after alignment. RANSAC
\cite{fischler1981random} can be used to get a more robust solution and
avoid the noise effect in building correspondences.  Since objects in
the gallery set are pre-aligned, the obtained rotation and translation
gives the 6-DOF pose of the input query object. 

\section{Experiments}\label{sec:results}

We use point cloud objects from the ModelNet40 dataset \cite{wu20153d}
in all experiments. For retrieval and pose estimation, we focus on four
object categories: airplane, chair, sofa, and car. For every object, the
dataset consists of 2048 points sampled from the surface. We use the
original train/test split. The experiments are divided into three parts
as discussed below. For R-PointHop, we use two hops and adjust the
energy threshold so as to have a 200-D point feature. The farthest point
sampling strategy is used after the first hop where the point cloud is
downsampled by half. 

\subsection{Object Registration with FR-PointHop}

After replacing the conventional R-PointHop attributes with the FPFH
descriptor, we see an improvement in all object registration tasks of
FR-PointHop over R-PointHop.  A random rotation and translation is
applied to the input object.  For the challenging case of partial object
registration (under the assumption that the partial reference point
cloud is available), we show the rotation and translation errors Table
\ref{tab:registration} in terms of the mean squared error (MSE), the
root mean squared error (RMSE) and the mean absolute error (MAE).  We
conduct performance benchmarking with SPA \cite{kadam2020unsupervised}
(SSL-based), FGR \cite{zhou2016fast} (FPFH-based), and PRNet
\cite{wang2019prnet}.  Clearly, the modified feature representation in
FR-PointHop is favorable even in presence of the reference object. 

\begin{table}[htbp]
\centering
\caption{Performance comparison of object registration.} \label{tab:registration}
\renewcommand\arraystretch{1.3}
\newcommand{\tabincell}[2]{\begin{tabular}{@{}#1@{}}#2\end{tabular}}
\begin{tabular}{c| c c c | c c c } \hline 
\textbf{} &\multicolumn{3}{|c|}{\textbf{Rotation error} (in degree)} &\multicolumn{3}{|c}{\textbf{Translation error}} \\ \hline
\bf Method & \tabincell{c}{\bf MSE}  &  \tabincell{c}{\bf RMSE}  
& \tabincell{c}{\bf MAE} & \tabincell{c}{\bf MSE}  &  \tabincell{c}{\bf RMSE} 
& \tabincell{c}{\bf MAE}  \\ \hline 
\tabincell{c} 
SPA \cite{kadam2020unsupervised}  & 229.09 & 15.13  & 8.22 & 0.0019 & 0.0435 & 0.0089   \\ \hline
FGR \cite{zhou2016fast}  & 126.29 & 11.24  & 2.83 & 0.0009 & 0.0300 & 0.0080   \\ \hline
PR-Net \cite{wang2019prnet} & 10.24 & 3.12 & 1.45 & 0.0003 & 0.0160 & 0.0100   \\ \hline
R-PointHop  & \underline{2.75} & \underline{1.66} & \underline{0.35} & \bf{0.0002} & \bf{0.0149} & \underline{0.0008}   \\ \hline
{\em FR-PointHop}  & \bf{2.68} & \bf{1.64} & \bf{0.33} & \bf{0.0002} & \bf{0.0149} & \bf{0.0007}   \\ \hline
\end{tabular}
\end{table}

\subsection{Object Retrieval with PCRP}

We use Precision@M and the chamfer distance as two evaluation metrics
for object retrieval. For every query object, we generate its ground
truth by rank-ordering objects in the gallery set based on the least
chamfer distance. The chamfer distance is calculated as the sum of
Euclidean distances of every point in one point cloud to its nearest
point in the other point cloud. The Precision@M is an average measure of
the number of top $M$ retrieved objects that match with the top $M$
objects from its ground truth. We expect a higher Precision@M score
while a lower chamfer distance for good retrieval methods.

Ten codewords are used in the VLAD implementation. we report
Precision@10 scores and Top-1 chamfer distances for two cases in Table
\ref{tab:retrieval}. First, the query object is aligned with those in
the gallery set. Second, a uniform random rotation and translation is
applied to the query object so that it has an arbitrary pose. We provide
comparisons with PointNet \cite{qi2017pointnet} (an exemplary deep
learning method), CORSAIR \cite{zhao2021corsair} (on similar lines to
our work, but supervised), PointHop \cite{zhang2020pointhop}
(SSL-based), and FPFH \cite{rusu2009fast}. For PointNet and PointHop,
globally pooled features are adopted for retrieval. For FPFH, we
aggregate point features using VLAD.  Furthermore, we replace our VLAD
aggregated feature with max pooling and report the performance
separately. 

Results in Table \ref{tab:retrieval} show the superiority of PCRP over
others. For PointNet and PointHop, we see a drop in performance in the
case of arbitrary poses. We expect similar performance for other methods
that do not take pose variations into account. Since CORSAIR, FPFH and
PCRP use pose invariant feature representations, their performance is
robust against pose variation. PCRP is the best among the three.
Moreover, aggregating local features with max pooling in PCRP degrades
the performance significantly, thereby justifying the inclusion of VLAD
in PCRP.

\begin{table}[htbp]
\centering
\caption{Comparison of point cloud retrieval performance.} \label{tab:retrieval}
\renewcommand\arraystretch{1.3}
\newcommand{\tabincell}[2]{\begin{tabular}{@{}#1@{}}#2\end{tabular}}
\begin{tabular}{c | c c | c c  } \hline 
& \multicolumn{2}{c|}{Pre-aligned objects} & \multicolumn{2}{c}{Arbitrary poses} \\ \hline
Method & \tabincell{c}{Precision@10 \\ (\%)}  &  \tabincell{c}{Top-1 Chamfer \\ distance}  
& \tabincell{c}{Precision@10 \\ (\%)} & \tabincell{c}{Top-1 Chamfer \\ distance}  \\ \hline 
\tabincell{c} 
PointNet \cite{qi2017pointnet} & 60.66 & 0.121 & 53.40 & 0.145   \\ \hline
PointHop \cite{zhang2020pointhop} & 58.23 & 0.129 & 19.71 & 0.211 \\  \hline
FPFH \cite{rusu2009fast} & 53.23 & 0.164 & 52.12 & 0.160  \\  \hline
CORSAIR \cite{zhao2021corsair} & \underline{61.28} & \underline{0.106} & \underline{61.24} & \bf{0.107}  \\  \hline
{\em PCRP (max pool)} & 43.23 & 0.147 & 41.89 & 0.131 \\  \hline
{\em PCRP (VLAD)} & \bf{63.23} & \bf{0.101} & \bf{63.07} & \underline{0.111}  \\  \hline
\end{tabular}
\end{table}

\subsection{Object Pose Estimation with PCRP}

We report mean and median rotation errors between the predicted and
ground truth pose for all four object classes in Table \ref{tab:pose}.
For performance benchmarking, we select ICP \cite{besl1992method} and
FGR \cite{zhou2016fast} among traditional methods for which we provide the template point cloud. For learning-based methods, we selected Chen \etal
\cite{chen2021equivariant} (supervised) and Li \etal
\cite{li2021leveraging} (self-supervised) two methods, which are based
on equivariant networks. Finally, CORSAIR \cite{zhao2021corsair} is also included. 

\begin{table}[htbp]
\centering
\caption{Mean and median rotation errors in degrees.} \label{tab:pose}
\renewcommand\arraystretch{1.3}
\newcommand{\tabincell}[2]{\begin{tabular}{@{}#1@{}}#2\end{tabular}}
\begin{tabular}{c | c c  c c  c c  c c} \hline 
 & \multicolumn{2}{c}{Chair} & \multicolumn{2}{c}{Airplane} & \multicolumn{2}{c}{Car} & \multicolumn{2}{c}{Sofa}\\ 
Method & \tabincell{c}{Mean}  &  \tabincell{c}{Median}  
& \tabincell{c}{Mean} & \tabincell{c}{Median}  &  \tabincell{c}{Mean} 
& \tabincell{c}{Median} & \tabincell{c}{Mean} & \tabincell{c}{Median} \\ \hline 
\tabincell{c} 
ICP \cite{besl1992method}  & 88.92 & 96.28  & 8.11 & \underline{1.22} & 22.76 & 2.94 & 39.00 & 9.69 \\ \hline
FGR \cite{zhou2016fast} & 22.10 & 6.04  & 6.84 & 3.33  & 18.44 & 2.69 & 9.97 & \underline{2.36} \\ \hline
CORSAIR \cite{zhao2021corsair}  & 13.99 & 4.58  & 8.09 & 3.43 & 12.09 & 2.13 & 9.12 & 3.24 \\ \hline
Chen \etal \cite{chen2021equivariant}  & \bf{8.56} & \bf{3.87}  & \bf{3.35} & \bf{1.12} & \bf{9.48} & \bf{1.85} & \bf{4.76} & \bf{1.56} \\ \hline
Li \etal \cite{li2021leveraging}  & \underline{7.05} & 4.55  & 23.09 & 1.66 & 17.24 & 2.13 & 8.87 & 3.22 \\ \hline
{\em PCRP} & 14.42 & \underline{4.24}  & \underline{2.98} & 1.65 & \underline{11.22} & \underline{2.11} & \underline{8.84} & 2.29 \\ \hline
\end{tabular}
\end{table}

From mean and median rotation errors, PCRP is significantly better than
ICP which is only good for local registration. It also outperforms FGR
and CORSAIR.  Its performance is slightly inferior to the method of Li
\etal \cite{li2021leveraging}. In all the cases, the mean rotation
errors are higher than the median error. We study the distribution of
rotation errors across all point clouds and observe that the higher
error is only due to a large registration error in only a few point
clouds. Actually, after plotting the CDF of the error for all point
clouds, more than 90\% of test samples have a rotation error less than 5
degrees. 

It is worthwhile to point out the advantages of PCRP. First, it combines
unsupervised feature learning with established non-learning-based pose
estimation using point correspondences. The training time is typically
less than 30 minutes in building the Saab kernels and the VLAD codebook.
The FR-PointHop model size is only 230kB along with 1.6MB to store the
VLAD features and the codebook.  In contrast, we find that the model
size of the exemplary works is 30MB for Chen \etal and 72MB for Li
\etal. This clearly highlights that PCRP would be favorable in resource
constrained occasions. 

Further investigation into failure cases reveals that most of them occur
when a similar matching point cloud is not available in the database for
retrieval. This is a bottleneck of PCRP. Under the assumption that a
similar object is present in the gallery set, it can estimate the pose
accurately. One way to filter out query samples automatically is to
compare the Chamfer distance to the best retrieved object. If the
Chamfer distance is above a certain threshold, it cannot be treated as a
reliable result. 

\section{Conclusion}\label{sec:conclusion}

An unsupervised feature learning method for point cloud retrieval and
pose estimation, called PCRP, was proposed in this paper. PCRP estimates
the 6-DOF pose comprising of rotation and translation of a point cloud
object using a similar pre-aligned object of the same object category.
It uses the R-PointHop method to extract point features from 3D point
cloud objects. The features are aggregated into a global descriptor
using VLAD and used to retrieve similar pre-aligned objects.  Finally,
the point features of the query and retrieved point clouds are used to
estimate the 3D pose of the query point cloud using registration. 

\bibliographystyle{unsrt}
\bibliography{refs}

\end{document}